\documentclass[sigconf]{acmart}
\renewcommand\footnotetextcopyrightpermission[1]{} 
\usepackage[linesnumbered, ruled, vlined]{algorithm2e}
\usepackage{algpseudocode}
\usepackage{makecell}
\usepackage{multirow}
\usepackage{gensymb}
\usepackage{amsmath}
\usepackage{booktabs,caption, threeparttable}
\graphicspath{ {images/} }

\def\BibTeX{{\rm B\kern-.05em{\sc i\kern-.025em b}\kern-.08emT\kern-.1667em\lower.7ex\hbox{E}\kern-.125emX}}
    
\begin{document}

%
\title{Predicting Unplanned Readmissions in the Intensive Care Unit: A Multimodality Evaluation}

%

\author{Eitam Sheetrit}
\authornote{Equal contribution}
\affiliation{%
  \institution{Microsoft}}
\email{eitams@microsoft.com}

\author{Menachem Brief}
\authornotemark[1]

\affiliation{%
  \institution{Microsoft}}
\email{t-mbrief@microsoft.com}

\author{Oren Elisha}
\affiliation{%
  \institution{Microsoft}}
\email{oren.elisha@microsoft.com}

%
\renewcommand{\shortauthors}{Sheetrit, Brief and Elisha}

%
\begin{abstract}
A hospital readmission is when a patient who was discharged from the hospital is admitted again for the same or related care within a certain period. Hospital readmissions are a significant problem in the healthcare domain, as they lead to increased hospitalization costs, decreased patient satisfaction, and increased risk of adverse outcomes such as infections, medication errors, and even death. The problem of hospital readmissions is particularly acute in intensive care units (ICUs), due to the severity of the patients' conditions, and the substantial risk of complications. Predicting Unplanned Readmissions in ICUs is a challenging task, as it involves analyzing different data modalities, such as static data, unstructured free text, sequences of diagnoses and procedures, and multivariate time-series. Here, we investigate the effectiveness of each data modality separately, then alongside with others, using state-of-the-art machine learning approaches in time-series analysis and natural language processing. Using our evaluation process, we are able to determine the contribution of each data modality, and for the first time in the context of readmission, establish a hierarchy of their predictive value. Additionally, we demonstrate the impact of Temporal Abstractions in enhancing the performance of time-series approaches to readmission prediction. Due to conflicting definitions in the literature, we also provide a clear definition of the term Unplanned Readmission to enhance reproducibility and consistency of future research and to prevent any potential misunderstandings that could result from diverse interpretations of the term.
Our experimental results on a large benchmark clinical data set show that Discharge Notes written by physicians, have better capabilities for readmission prediction than all other modalities. 
\end{abstract}

%

%
\maketitle

\section{Introduction}
\label{sec:introduction}

The process of discharging patients from hospitals is complicated and presents many difficulties, with more than 35 million hospital discharges taking place each year in the United States alone~\cite{1-CDC2018}. One such difficulty is hospital readmission. 
Hospital readmission refers to the occurrence of a patient being admitted to the hospital again for the same or related care within a certain period, often within 30 days, after being discharged. Hospital readmissions cause a significant financial burden, an avoidable waste of medical resources, and can also put a strain on the healthcare system~\cite{3-lin2019analysis}. The cost of unplanned readmissions in the United States is estimated at 15 to 20 billion dollars annually~\cite{2-jencks2009rehospitalizations}.

The creation of the Hospital Readmissions Reduction Program (HRRP) in 2010 by the Affordable Care Act (ACA) aimed to tackle the problem of unplanned hospital readmissions. The program imposes penalties on general acute-care hospitals whose rates of readmissions within 30 days exceed a maximal allowed readmission rate per clinical category. According to data released by the Centers for Medicare and Medicaid Services (CMS), since the beginning of this program, 93\% of the 3,139 general acute hospitals subject to HRRP evaluation have been penalized at least once, with a total of 2.5 billion dollars of penalties for readmissions~\cite{4-10YearsOfHospitalReadmissionsPenalties}.

When focusing on intensive care unit (ICU) readmissions, severe health consequences are added to the economic implications. Patients who are readmitted to the ICU after being transferred to the medical floor have a significantly higher mortality rate (ranging from 21\% to 40\%) compared to those who are not readmitted (with a mortality rate ranging from 3.6\% to 8.4\%)~\cite{5-mcneill2020impact}. 

Reducing ICU readmission rates is an important goal for healthcare providers and policymakers, as it can improve patient outcomes, reduce length of stay that ultimately effects the availability of ICU beds, and reduce healthcare costs. To achieve this goal, hospitals need to identify patients with a high risk of ICU readmission. An accurate decision support system can assist physicians in identifying such patients, therefore reducing their risk for readmissions by prolonging their ICU stay, improving their follow-up care, or providing a comprehensive discharge plan. 

The usual and expected candidates for readmission are patients that were hospitalized within the past year~\cite{Mudge2011RecurrentRI}, often suffering from chronic conditions, and thus can be easily identified by physicians. As the aim of predicting unplanned readmissions is to detect instances of unexpected and unforeseen readmissions, it is important to identify and exclude such patients from the analysis, to avoid mixing expected and unplanned readmissions. Throughout the rest of this paper we use \textit{readmission} synonymously with this excluding definition of \textit{unplanned readmissions}.

Every ICU stay involves a vast amount of heterogeneous data: (1) static, such as patients demographics, (2) sequential, such as diagnoses and prescriptions, (3) unstructured, high-dimensional and sparse information such as clinical discharge notes, and (4) dynamic multivariate time-series data, such as lab tests and near bedside measurements, with problems common in clinical medicine, of missing data, different sampling frequencies, and random noise.
The multimodal nature of the data makes predicting readmission a difficult task.

\begin{figure*}
  \includegraphics[width=\textwidth]{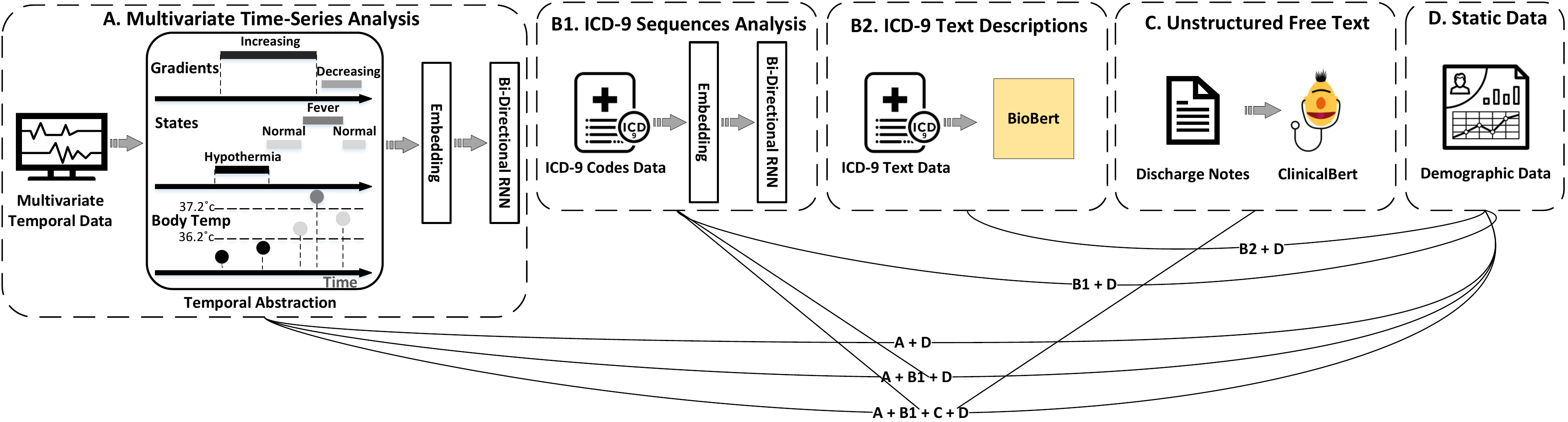}
  \caption{Overview of the proposed approach. It's comprised four components, each for every data modality. Except for (D), every component is a standalone method for readmission prediction. The lines at the bottom describe multimodality methods combinations for the given task.}
  \label{fig:fig1-overview}
\end{figure*}

In this paper, we investigate each data modality, and its influence on unplanned ICU readmission prediction. 
Figure~\ref{fig:fig1-overview} illustrates our proposed approach. It is comprised of four building blocks that represent every data modality of input that is being processed using three communicative models: (A) raw, time-stamped multivariate temporal data collected by various sensors, which are abstracted into several interval-based abstract concepts, each belonging to one of two abstraction types, as we explain in Section~\ref{subsec:time-series}; (B) diagnoses and procedures performed on patients, represented as sequences of relevant ICD-9 (International Classification of Diseases, Ninth Revision) codes (B1), or ICD-9 textual descriptions (B2); (C) discharge notes, which are unstructured free text, written by physicians when releasing patients from the ICU; and (D) Static data on patients that includes their demographics, insurance type, etc. 
Besides (D), every data modality is examined separately, and in combination with others, as illustrated at the bottom of Figure~\ref{fig:fig1-overview}.

Our contributions in this study are as follows: (1) Demonstrating the effectiveness of each data modality separately, and every combination of them, on the prediction results. To the best of our knowledge, this is the first time multimodality analysis is being conducted in the task of ICU readmission prediction. (2) Combining different temporal-abstraction types in addition to domain knowledge, to better capture the temporal behavior of the data, thus improving the performance of temporal modalities. (3) Redefining the Unplanned Readmission problem, and creating a specific subset of patients to use in future works, enhancing reproducibility and cross study comparisons. 

The rest of the paper is structured as follows. The next section provides background on time-series, temporal abstraction and on natural language processing. 
Section~\ref{sec:related-work} reviews related work.
Section~\ref{sec:data} describes the process of creating the dataset used in the evaluation. 
Section~\ref{sec:evaluation} presents an experimental evaluation and all methods involved, in the ICU domain.
Section~\ref{sec:results} provides the evaluation's results.
Section~\ref{sec:Discussion} discusses our insights from the results and concludes the paper.

\section{Background}
\label{sec:background}

\subsection{Time-Series and Temporal Abstraction}
\label{subsec:time-series}
Time series are sequences of discrete or continuous real-valued elements collected over time, such as network monitoring data, stock market data, and patient data stored in medical records. The entity (e.g., a patient) is sampled 
at fixed or varying time periods. The data may later be represented as time points for some variables (e.g., a body temperature of $ 38\degree C $ at 19:30:00 on 01/03/23), or as time intervals (e.g., a fever lasting from 01/03/23 to 02/03/23). Often, the entity's collected data are multivariate, originating from multiple sensors that sample the data in frequencies that vary between sensors and within each sensor. Calculating similarities between different sets of such multivariate time series is challenging, as we need to deal with different sampling frequencies, lack of synchronization between sensors, missing data, and random changes in entities' behavior over time. 

Basing the analysis on interval-based temporal \textit{abstract} concepts and domain knowledge (e.g., \textit{a fever} over eight hours), rather than the original time-stamped \textit{raw} data, might be helpful in reducing random noise, avoiding problems resulting from sampling the data at different frequencies, summarizing large numbers of data in a succinct fashion~\cite{55-goldstein2016automated}, and helping algorithms to cope with missing data~\cite{3-moskovitch2015fast}.

\textit{Temporal abstraction} (TA) is the task of representing time-stamped raw data as a set of time intervals, typically at a higher level of abstraction~\cite{15-shahar1998timing,56-combi1997temporal}. Given a set of time-stamped data, external events, and sometimes also abstraction goals (which provide additional contexts for the TA process), TA produces a set of interval-based abstractions of the data that represent past and present states and trends that are relevant for the given set of goals. 
The interval-based abstractions are sensitive to the \textit{context} in which each datum appears~\cite{57-shahar1998dynamic}, such as a blood-glucose measurement taken within the temporal span of the effect of a short-acting Insulin administration, or weight measurements of infants versus those of adults.

TA, which in most cases includes some form of interpolation~\cite{16-shahar1999knowledge}, is often helpful when dealing with data that were sampled at different frequencies or that are characterized by multiple missing values due to its smoothing effect on the generated abstractions. Temporal abstraction techniques usually include a discretization module as a pre-processing step, which discretizes raw data into a small number of possible ordinal categories (e.g., 'low', 'normal', 'high'), by determining cut-offs values. The discretization can be performed either by automatic unsupervised techniques~\cite{17-dougherty1995supervised,20-morchen2005optimizing}, supervised methods~\cite{8-moskovitch2015classification}, or by using knowledge acquired from a domain expert~\cite{9-shahar1997framework}, allowing us to embed human knowledge in the machine learning process.

\subsection{Natural Language Processing}
\label{subsec:nlp}
Natural Language Processing (NLP) is a sub-field of machine learning that focuses on allowing machines to understand and generate human language, as well as perform other downstream language-based tasks. 

Classic NLP methods have typically involved building rule-based systems that attempted to mimic human language processing. These systems relied on hand-crafted rules and linguistic analysis to identify and extract features from text. While these methods were effective to some extent, they were limited in their ability to handle the complexity and variability of natural language. Further advancements included explicit probabilistic modeling of language (LMs) using tools such as Markov chains and bag-of-words approaches. With the rise in popularity of neural networks, variations of recurrent neural networks (RNNs) have also been used with increasing success~\cite{Jurafsky2000SpeechAL}. 

In recent years, the transformer architecture~\cite{Vaswani2017AttentionIA} has become the state-of-the-art approach in NLP. Transformer models are deep neural networks, usually trained on massive amounts of text data, which use a technique called self-attention to learn contextual relationships between words. This allows them to generate more natural and fluent language, as well as produce state-of-the-art results for most NLP tasks, including language translation, sentiment analysis, and question answering. 

 Following the introduction of transformer based large language models (LLMs), and BERT~\cite{Devlin2019} in particular, tasks involving deep understanding of natural language have shown dramatic improvement. Furthermore, pre-training BERT architectures on medical text has shown that LLMs can achieve even better performance on downstream clinical tasks~\cite{Gu2020DomainSpecificLM}. More specifically, Gu et al. have shown that their model, BioBert achieves state-of-the-art performance on multiple clinical tasks. 

\section{Related Work}
\label{sec:related-work}
The combination of machine learning in general, and deep learning in particular, with data from electronic health records (EHR) has been proven to be useful for various clinical applications, including readmission~\cite{Morgan2019AssessmentOM}. While useful, EHR data is often very complex in structure, spanning multiple modalities, namely, tabular data, time series, and free text. As such, many works have focused on the EHR records from the perspective of either time-series methods or NLP methods. Multiple works have demonstrated that deep learning yields better results than classic machine learning algorithms~\cite{Wang2018PredictingHR, Pham2016, Lin2018AnalysisAP}. Nonetheless, various studies in the past 
~\cite{Caruana2015IntelligibleMF, Sushmita2016Predicting3R} have utilized linear and ensemble models, combined with feature engineering, to achieve varying degrees of success in classification of readmission.      

Proper utilization of the temporal data available in EHRs is a non-trivial task. EHR data is characterized by irregular sampling frequencies, both across measurement types as well as internally. This leads to many missing values when treating the data as a multivariate time series. 
Various works have bench-marked a variety of time series architectures~\cite{Barbieri2019BenchmarkingDL, Lin2018AnalysisAP, Pham2016}, predominantly favoring variations of RNNs, including combinations of RNNs with convolutional neural networks (CNNs). While multiple works, have demonstrated success using temporal methods, to the best of our knowledge only~\cite{Lin2018AnalysisAP} have conducted ablation studies to directly assess the predictive value of EHR temporal data. All other works have incorporated additional data modalities, most importantly tabular data describing doctors' diagnoses, making it difficult to isolate the contribution of each component. 

In contrast to the temporal data, utilizing the free text available has been much more straightforward. Huang et al~\cite{Huang2019ClinicalBERTMC} have shown that pre-training a BERT based model, ClinicalBERT, on EHR doctor notes, specifically discharge summaries, followed by fine-tuning, can yield better results than other NLP models. To overcome limits on input size to the model, a sentence aggregation method was employed. Further improvements haven been shown to be possible~\cite{Golmaei2021DeepNoteGNNPH} using note embeddings to create a similarity based patient graph, and then using graph neural networks. Currently, these type of models are the state-of-the-art in readmission prediction.

A major issue that arises when comparing the various works in the field is that prior to the introduction of MIMIC-III~\cite{johnson2016mimic}, and even afterwards, many works focus on proprietary datasets, making their results unreproducible. More so, every selection of a subset of patients yields different results, even within a single dataset (for example, results reported on ClinicalBERT in~\cite{Huang2019ClinicalBERTMC} vs.~\cite{Golmaei2021DeepNoteGNNPH}). This makes direct comparison of results reported in different studies highly problematic. 
In this study we aim to solve this last problem, by presenting a standardized subset of patient admissions for future evaluation of Unplanned Readmission methods.

\begin{table}
  \caption{Readmission related knowledge base}
  \label{tab:readmission_kb}
  \begin{tabular}{lccr}
    \toprule
    \thead{Clinical \\ Concept} & \thead{"Normal" State \\ Values} & \thead{Gradient/Trend \\ sigValues} & \thead{Trend \\ tStable} \\
    \midrule 
    Chloride & 96-106 mEq/L & $\Delta > 5$ & 36 hours\\
    Creatinine & 0.6-1.3 mg/dL & $\Delta > $ 0.2 & 36 hours\\
    Glucose & 70-100 mg/dL & $\Delta > $ 10 & 36 hours\\
    Hemoglobin & 11-18 g/dL & $\Delta > $ 2 & 36 hours\\
    PCO2 & 38-42 mm Hg & $\Delta > $ 2 & 36 hours\\
    PH & 7.34-7.45 pH & $\Delta > $ 0.05 & 36 hours\\
    Phosphate & 2.4-4.1 mg/dL & $\Delta > $ 0.5 & 36 hours\\
    PLT & 150-400 $x10^9$/L & $\Delta > $ 50 & 36 hours\\
    PO2 & 75-100 torr & $\Delta > $ 10 & 36 hours\\
    Urea & 10-20 mg/dL & $\Delta > $ 5 & 36 hours\\
    Sodium & 135-145 mEq/L & $\Delta > $ 5 & 36 hours\\
    WBC & 4.5-10 $x10^9$/L & $\Delta > $ 1 & 36 hours\\
    \midrule
    Body Temp & $36.2-37.2^\circ$C & $\Delta > $ 0.5 & 2 hours\\
    Glasgow CS \tnote{*} & 8-12 & $\Delta > $ 2 & 2 hours\\
    Mean Pressure & 65-80 & $\Delta > $ 5 & 1 hour\\
    Heart-Rate & 60-80 bpm & $\Delta > $ 10 & 1 hour\\
    Respiratory-Rate & 7-14 breath/min & $\Delta > $ 3 & 1 hour\\
  \bottomrule
\end{tabular}
\begin{tablenotes}
     \item *Glasgow Coma Scale states are mild, moderate and severe
   \end{tablenotes}
\end{table}

\section{Data Collection}
\label{sec:data}
Our evaluation uses the MIMIC-III dataset~\cite{johnson2016mimic}, which is comprised of de-identified health-related data associated with over 40,000 patients who stayed in intensive care units (ICUs) at Beth Israel Deaconess Medical Center between 2001 and 2012. The database includes information such as demographics, vital sign measurements made at the bedside, laboratory test results, doctor notes, and more. It consists of 26 tables that record every piece of data for the patient's admission to the hospital. 

\subsection{Data Modalities}
\label{subsec:data_mods}
We view the data in MIMIC-III as being composed of three main data modalities, temporal, tabular and textual.

Following~\cite{Barbieri2019BenchmarkingDL, Sheetrit2019TemporalPP}, we identified 17 clinical concepts from the temporal data in MIMIC-III that are related to readmission. These can be divided into two groups: (1) lab tests (i.e., WBC, Hemoglobin, etc.), sampled with low frequency (daily, not hourly); (2) chart items (body temperature, blood pressure, heart rate, etc.), sampled with high frequency (hourly/minutely). Then we established knowledge-based state and gradient abstraction definitions, using guidelines which are described in Table~\ref{tab:readmission_kb}. The definitions were used for the abstraction of the raw data into three to six states, and three gradients (increasing, decreasing or stable values). We will refer to this data as \textit{Charts}. 

Throughout a patient's stay, hospital staff also record all doctor diagnoses, including pre-existing medical conditions, as well as the procedures a patient went through. This data is recorded using standardized ICD-9 codes. While the data is provided in a table using the codes, a textual description is also available for each code. This data modality is pseudo-tabular, since it can be viewed as a low frequency time series or it can be flattened into table form. We will refer to this data as \textit{ICD-9}. 

Finally, a large body of notes is included for each stay. Following~\cite{Huang2019ClinicalBERTMC}, we chose to focus on the discharge notes. These contain a brief summary of the patients stay as well as any other relevant information the doctor chooses to include, mostly in unstructured free-form text. We will refer to this data as \textit{Notes}.  

\begin{table}
 \caption{Distribution of patients in the dataset. Number of readmissions is colored in black.}
  \centering
  \begin{tabular}{lll}
    \toprule
    \textbf{Gender}     & \textbf{Ages 18-65}     & \textbf{Ages $>$ 65} \\
    \midrule
    Male & \textbf{485}/\color{gray}\textbf{4268}  & \textbf{531}/\color{gray}\textbf{3569}     \\
    Female     & \textbf{303}/\color{gray}\textbf{2659} & \textbf{433}/\color{gray}\textbf{3176}      \\
    \bottomrule
  \end{tabular}
	\label{table:Distribution_of_patients}
\end{table}

\subsection{Dataset Creation}
\label{subsec:data_creation} 

We examined the data of adult patients (18 years old or older), that were admitted to the ICU and remained there for at least one day and no more than 30 days (to remove outliers). We labeled our data as positive if a patient has returned to the ICU within the 30 days following their discharge, and negative otherwise. In order to properly evaluate the contribution of each data modality we removed admissions that did not have at least one of each lab tests measurement, and at least five of each chart items measurements. We also removed patients that died during the stay or throughout the 30 following days, since they clearly cannot be readmitted. Finally, we removed all but the first ICU stay of each patient in every year. This is important for two reasons: many hospital admissions may include multiple ICU stays before being discharged, and the usefulness of predicting readmission for patients that have had many recent readmissions is limited, as explained in~\ref{sec:introduction}. Those conditions resulted in a total of 14,837 patients across 15,424 ICU stays: 1,752 readmissions (11.3\%) and 13,672 that were not readmitted. These represent the entirety of patients relevant to Unplanned Readmissions in MIMIC-III, per our definition. The demographic distribution of the patients can be seen in Table~\ref{table:Distribution_of_patients}, where black and gray indicate the number of positive and negative labels respectively.

We then proceeded to divide the patients into five folds randomly, leaving (20\%) as a test set for each fold. The split was done by patient and not by ICU stay, to avoid data leakage across a patient's various stays.  The splits were done in a stratified way, to ensure a similar distribution of labels in all five test sets. These splits were used for all of our experiments, and the reported results are aggregations across all folds.

\section{Evaluation Methods}
\label{sec:evaluation}

Due to the multiple modalities, our experiments were split so that an ablation study regarding the predictive value of each modality could be conducted properly. All models and modalities were trained using the same training and evaluation loop as described in Appendix~\ref{sec:training-details}. Our code was all written in Python, with deep learning components using PyTorch~\cite{Paszke2019PyTorchAI}, and Random Forest and evaluation metrics using Scikit-learn~\cite{scikit-learn}. 

\subsection{Key Architectures}
\label{subsec:key-arcs}
The two main architectures used throughout our experiments are BERT~\cite{Devlin2019} and a Gated Recurrent Network (GRU)~\cite{Chung2014EmpiricalEO} based bidirectional RNN network (BIRNN). 

For the BERT models we used the standard architecture, available from Hugging Face's Transformers library~\cite{wolf-etal-2020-transformers}, followed by a fully connected layer for classification. 

\begin{figure}
  \includegraphics[width=0.5\textwidth]{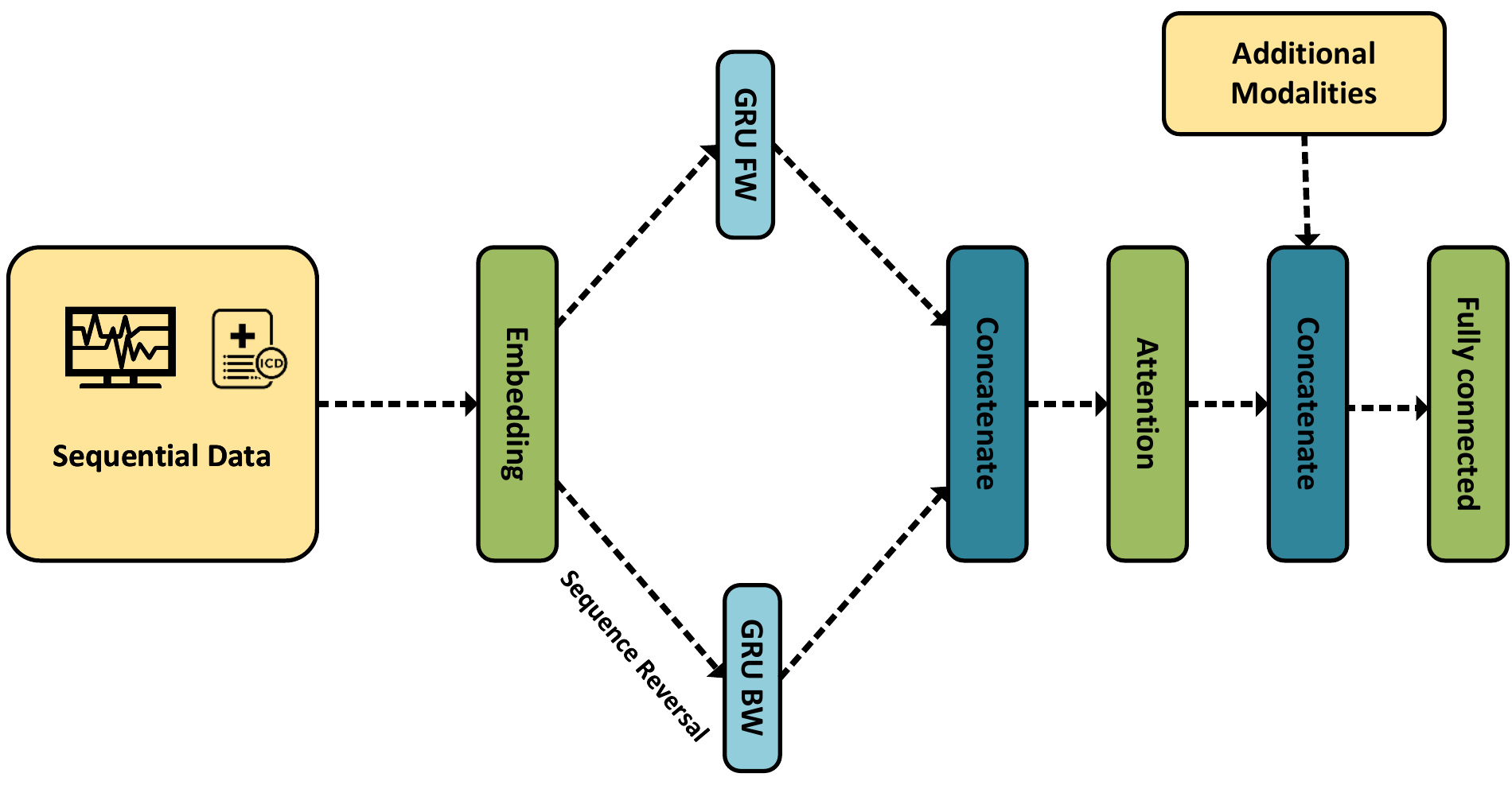}
  \caption{BIRNN architecture.}
  \label{fig:fig2-BIRNN}
\end{figure}

The BIRNN's architecture is composed of four main steps: an Embedding layer, two GRUs, an attention layer, and a fully connected layer. The embedding layer takes a vector from $\mathbb{R}^L$ and returns a $ \mathbb{R}^{L \times \lfloor \sqrt[4]{D} \rfloor}$ matrix, where $L$ is the maximal sequence length and $D$ is the number of unique values in the input. The embeddings are then passed to two GRUs, with the second one receiving a reversal (in the time dimension) of the input. The output of both GRUs is then concatenated together, and passed to the attention layer, providing a weighted version of its input. Finally, the previous output is concatenated to any additional modality if necessary, and a fully connected layer reduces each sample to a single value representing the score. A visual representation of the architecture can be seen in Figure~\ref{fig:fig2-BIRNN}. 

\label{subsubsec:BIRNN}

\subsection{Charts}
\label{subsec:exp-charts}
Being the most complicated data modality, we experimented with multiple data preparation methods. We compared each method from~\ref{subsubsec:exp-discretization} both on its own, as well as combined with the method from~\ref{subsubsec:exp-TAs}. We also performed a one hot encoding of the patient's demographics (\textit{demographics 1-hot}) and concatenated it to the input of each model trained on \textit{Charts}.\\
\subsubsection{Discretization}
\label{subsubsec:exp-discretization}
As our baseline, we used the method described by~\cite{Barbieri2019BenchmarkingDL}. We discretized the raw charts data into severity bins based on~\cite{Johnson2013, Sheetrit2019TemporalPP}, followed by flattening all measurements into a temporally ordered vector, i.e. a univariate time series. An illustration of this can be seen in Figure~\ref{fig:fig1-overview}.A, Temporal Abstraction, e.g. \textit{Body Temperature} below $ 36.2\degree C $ is categorized as "Hypothermia", between $ 36.2\degree C $ and $ 37.2\degree C $ as "Normal" and above $ 37.2\degree C $ as "Fever", and so on for each measurement, at each time step, using guidelines which are described in Table~\ref{tab:readmission_kb}. This vector was then one hot encoded and padded to the maximal length (\textit{Charts 1-hot}). \\

Our next method was to use a simple temporal-linear interpolation before the discretization, as described in Equation~\ref{eq-interpolate}:
\begin{equation} 
\label{eq-interpolate} 
x_i^{(\tau_t)} = x_i^{(\tau_{t+1})} \cdot \frac{\tau_{t+1} - \tau_t}{\tau_{t+1} - \tau_{t-1}} + x_i^{(\tau_{t-1})} \cdot \frac{\tau_{t} - \tau_{t-1}}{\tau_{t+1} - \tau_{t-1}},
\end{equation} 
where $x_i^{(\tau_{t-1})}, \; x_i^{(\tau_{t+1})}$ are two consecutive known data points of measurement $i$, and there is at least one other measurement that is known at time $\tau_t$. We then performed the same discretization as described above, except that instead of flattening, we left the data as a multivariate time series (\textit{Charts interpolated}). Note that the number of observations in both methods is the same, but now $x^{(\tau_{t})}$ is a 17-dimensional vector, rather than a single data point. \\
\subsubsection{Temporal Abstraction}
\label{subsubsec:exp-TAs}
Similar to what is described in~\ref{subsec:time-series}, we created TAs, describing the direction of change in the data. We will refer to these as \textit{gradients}. We used a simple method to characterize subsequent gaps, given as follows:
\begin{equation} 
\label{eq-simple_grad} 
gradient(x_i^{(\tau_t)}) = \begin{cases}

Increasing & x_i^{(\tau_t)} > x_i^{(\tau_{t-1})}\\
Decreasing & x_i^{(\tau_t)} < x_i^{(\tau_{t-1})}\\
Stable & x_i^{(\tau_t)} = x_i^{(\tau_{t-1})}\\
\end{cases}.
\end{equation} 

\subsubsection{Time series models}
\label{subsubsec:exp-ts_models}
We trained three different model architectures and selected the best architecture for the rest of the evaluations. The three we selected to compare were a BIRNN model based on the architecture described in~\ref{subsec:key-arcs}, an MLSTM-FCN model based on the architecture proposed in~\cite{Karim2018MultivariateLF}, and a TimeTransformer model based on the architecture proposed in~\cite{Wu2020DeepTM}. To make the comparison fair, we evaluated all three architectures on \textit{Charts 1-hot interpolated}. 

\subsection{ICD-9}
\label{subsec:exp-ICD9}
We prepared the data in two different ways. \\
In the first method, we took all the ICD-9 codes and flattened them into a single vector, essentially a one hot encoding. As each admission may have a varying number of codes, we padded all vectors to the maximal size (\textit{ICD-9 1-hot}).\\
In the second method, we simply concatenated the textual descriptions, separated by commas, instead of the codes (\textit{ICD-9 text}). We added a textual description of the patient's demographics as well as the text \textit{"Procedures patient went through and doctor's diagnoses:"} at the beginning. \\
As our baseline we used a random forest (RF) model, due to the pseudo-tabular nature of the data. Its input was \textit{ICD-9 1-hot} and \textit{demographics 1-hot} concatenated together. We then trained a BIRNN model, as described in ~\ref{subsec:key-arcs}. We elected to use a recurrent model over a linear model due to the sequential nature of the data, i.e. diagnoses are provided throughout the ICU stay, and ordered by time. Finally, we trained a BioBERT model~\cite{Gu2020DomainSpecificLM} using the pretrained weights Gu et al. released to fine tune a model with \textit{ICD-9 text} as its input.

\subsection{Notes}
\label{subsec:exp-notes}
We used the methods described in~\cite{Huang2019ClinicalBERTMC} to pre-process the discharge notes. This includes text cleaning, and tokenization. We also followed their suggested method for dealing with the note size, by dividing each note into $n$ chunks and recombining them by performing their suggested aggregation (Equation~\ref{eq-agg_notes}) on the model's output.
\begin{equation} 
\label{eq-agg_notes} 
P(readmission|h_{patient}) = \frac{P^n_{max} + P^n_{mean}\frac{n}{2}}{1 + \frac{n}{2}}.
\end{equation} 
We used the pre-trained model weights that they have published and perform fine-tuning for readmission classification. It is important to note that while the model may have been pre-trained using notes from some fold's test set, we do not see this as a data leakage issue. Our rationale is that a model can always be pre-trained with new discharge notes that do not yet have a label, similar to any other unannotated data used for unsupervised tasks. 

\subsubsection{Multi-modalities}
\label{subsubsec:exp-multi}
For completeness we also trained two multi-modal models. The first model was a combination of the best \textit{Charts} model with the BIRNN \textit{ICD-9} model. The second model was a combination of the previous model with the ClinicalBERT \textit{Notes} model. We combined the individual models into a single multi-modality model using the method described in~\ref{subsubsec:BIRNN}.

\begin{table*}[t]
\caption{Time series architectures comparison}
\label{tab:ts-architectures}
\begin{tabular}{l|lllll}
\hline
Architecture & AUROC               & F1                           & AUPRC               & Precision                   & Recall                       \\ \hline
BIRNN        & 0.6235 $\pm$ 0.0144 & \textbf{0.2517 $\pm$ 0.0126} & 0.1598 $\pm$ 0.0105 & \textbf{0.1635 $\pm$ 0.016} & 0.5604 $\pm$ 0.0525          \\
MLSTM-FCN    & 0.6188 $\pm$ 0.0135 & 0.2433 $\pm$ 0.0116          & 0.1509 $\pm$ 0.0116 & 0.1567 $\pm$ 0.0158         & \textbf{0.5853 $\pm$ 0.1528} \\
TimeTransformer & \textbf{0.6248 $\pm$ 0.0126} & 0.2484 $\pm$ 0.0128 & \textbf{0.1663 $\pm$ 0.0133} & 0.1618 $\pm$ 0.0151 & 0.5503 $\pm$ 0.0728 \\ \cline{1-6}
\end{tabular}
\end{table*}

\begin{table*}[t]
\caption{Charts data preparation comparison}
\label{tab:ts-prep}
\begin{threeparttable}
\begin{tabular}{l|lllll}
\hline
Method & AUROC & F1 & AUPRC & Precision & Recall \\ \hline
1-hot        & 0.6074 $\pm$ 0.0187 & 0.2429 $\pm$ 0.0105 & 0.1629 $\pm$ 0.006  & 0.1541 $\pm$ 0.0123 & \textbf{0.5936 $\pm$ 0.1029} \\
Interpolated & 0.6235 $\pm$ 0.0144 & 0.2517 $\pm$ 0.0126 & 0.1598 $\pm$ 0.0105 & 0.1635 $\pm$ 0.016  & 0.5604 $\pm$ 0.0525          \\ \hline
1-hot + gradients  (A + D)      & 0.6255 $\pm$ 0.0181        & \textbf{0.2573 $\pm$ 0.0164} & \textbf{0.1716 $\pm$ 0.0145} & \textbf{0.177 $\pm$ 0.0259} & 0.4968 $\pm$ 0.09  \\
Interpolated + gradients & \textbf{0.63 $\pm$ 0.0245} & 0.2524 $\pm$ 0.014           & 0.1684 $\pm$ 0.0171          & 0.1703 $\pm$ 0.016          & 0.5021 $\pm$ 0.081 \\ \hline
\end{tabular}
\end{threeparttable}
\end{table*}

\begin{table*}[t]
\caption{Full cross modality comparison}
\label{tab:full-results}
\begin{tabular}{l|lllll}
\hline
Method                      & AUROC               & F1                  & AUPRC               & Precision                    & Recall                     \\ \hline
1. ClinicalBERT (C) &
  \textbf{0.7522 $\pm$ 0.0241} &
  \textbf{0.3672 $\pm$ 0.0232} &
  \textbf{0.2988 $\pm$ 0.0278} &
  \textbf{0.2686 $\pm$ 0.0246} &
  0.5836 $\pm$ 0.0085 \\ \hline
2. ICD-9 RF &
  0.6873 $\pm$ 0.013 &
  0.293 $\pm$ 0.0163 &
  0.2243 $\pm$ 0.014 &
  {\color[HTML]{656565} \textbf{0.2384 $\pm$ 0.0344}} &
  0.413 $\pm$ 0.1022 \\
3. ICD-9 1-hot (B1 + D) &
  0.6932 $\pm$ 0.0101 &
  0.3113 $\pm$ 0.0082 &
  0.2285 $\pm$ 0.0139 &
  0.2228 $\pm$ 0.0119 &
  {\color[HTML]{656565} \textbf{0.52 $\pm$ 0.0329}} \\
4. ICD-9 Text (B2) &
  {\color[HTML]{656565} \textbf{0.6997 $\pm$ 0.0278}} &
  {\color[HTML]{656565} \textbf{0.3168 $\pm$ 0.0272}} &
  {\color[HTML]{656565} \textbf{0.2306 $\pm$ 0.0263}} &
  0.2343 $\pm$ 0.0281 &
  0.5052 $\pm$ 0.0802 \\ \hline
5. A + D &
  0.6255 $\pm$ 0.0181 &
  0.2573 $\pm$ 0.0164 &
  0.1716 $\pm$ 0.0145 &
  0.177 $\pm$ 0.0259 &
  0.4968 $\pm$ 0.09 \\
6. A + B1 + D &
  0.7042 $\pm$ 0.006 &
  0.3182 $\pm$ 0.0149 &
  0.236 $\pm$ 0.0096 &
  0.235 $\pm$ 0.0225 &
  0.502 $\pm$ 0.081 \\
7. A + B2 + D &
  0. 6911 $\pm$ 0.0166 &
  0.3074 $\pm$ 0.0234 &
  0.233 $\pm$ 0.0147 &
  0.224 $\pm$ 0.0326 &
  \textbf{0.7476 $\pm$ 0.1103} \\
8. A + B1 + C + D &
  0.7492 $\pm$ 0.0161 &
  0.3584 $\pm$ 0.0144 &
  0.293 $\pm$ 0.0161 &
  0.2672 $\pm$ 0.0109 &
  0.5461 $\pm$ 0.043 \\ \hline
\end{tabular}
\end{table*}

\section{Results}
\label{sec:results}

Throughout our evaluation we used five metrics to determine the quality of a model's predictions: area under the receiver operating characteristic curve (AUROC), F1 score, area under the precision recall curve (AUPRC), precision, and recall. While AUROC and AUPRC take the raw outputs of the model as input, the others were calculated by using an optimal threshold. Since some of these metrics are not directly correlated, i.e., a model may outperform another in AUROC but not in F1, we say a model is conclusively better than another if it receives a higher score in at least three metrics.\\
As can be seen in Table~\ref{tab:ts-architectures}, we found no conclusive winner between the BIRNN and the TimeTransformer architectures, and opted for the BIRNN architecture only because of its similarity to previous works. Both clearly outperformed the MLSTM-FCN. \\
In the comparison of four different \textit{Charts} data preparation methods, as shown in Table~\ref{tab:ts-prep}, we found that \textit{Charts interpolated} yielded significantly better results than \textit{Charts 1-hot}. This result seems to highlight the power of simple data imputation methods when dealing with missing values. Surprisingly, while adding \textit{gradients} conclusively improved both methods, \textit{Charts 1-hot} combined with \textit{gradients} emerged as the best \textit{Charts} model, outperforming \textit{Charts interpolated} with \textit{gradients}. The additional information gained when including \textit{gradients} seems to yield an improved effect when compared to interpolating. We believe this may be due to the improved ability to capture the relationship between states without providing "artificial" data. \\
Deep learning methods have proven to be conclusively better than classic ML methods for \textit{ICD-9}, as shown in Table~\ref{tab:full-results}. Results colored in gray signify that they are the best when using only \textit{ICD-9}. It is interesting to observe that using the full textual description \textit{ICD-9 text} was conclusively better than its one-hot encoded counterpart. We think that this stems from the pretraining knowledge incorporated within BioBERT, allowing the model a deeper semantic understanding. Following the same line of reasoning, we were not surprised that the ClinicalBERT model, trained on \textit{Notes}, achieved the best scores across the board, and by a significant margin. Furthermore, \textit{Notes} is written in formats that are intended to be informative, allowing for an easier understanding of an admission, which we believe may assist the model. \\
As expected, the multi-modality model that included both \textit{ICD-9} and \textit{Charts} (Table~\ref{tab:full-results}.6), was a conclusive improvement. It is easy to observe that each of these two modalities contain distinctly different information, thus the combination allowed for a richer representation of an admission. Nonetheless, the overall contribution of adding \textit{Charts} was relatively small.
On the other hand, the second multi-modality model (Table~\ref{tab:full-results}.8) demonstrated that combining all three together yields worse results that simply using \textit{Notes} on its own. A possible explanation for this is that unlike the other two modalities, \textit{Notes} may already include all the important information from other modalities, pre-filtered by doctors.

\section{Discussion and Conclusions}
\label{sec:Discussion}
One of the main problems when dealing with clinical data is the vast amount of heterogeneous raw data originating from various sources. It includes static demographic data on patients, sequences of diagnoses and procedures, doctor notes as free unstructured text, and multivariate time-series data that in most cases is sampled irregularly (in time), with various sampling frequencies and missing values.
In this paper, we demonstrated the effectiveness of every type of data on the challenging task of predicting patients readmissions. We defined the unplanned readmission problem in a strict way: we only took into account unplanned readmissions, i.e., ICU hospitalizations of patients who had not been hospitalized in an ICU during the previous year. We did so to avoid cases of patients who have chronic conditions and a high likelihood of being readmitted for the same issue in a short period, patients who are the usual candidates for readmission. It is important to identify and avoid including such patients in our evaluation, as they are easily recognizable by physicians, and our goal was to identify unplanned and unexpected cases of readmission.

While previous research on Readmission prediction usually focused on one or two data modalities, in this paper we investigated the prediction performance using every modality separately, and using different combinations of them, as each data modality may capture different aspects of a patient's hospitalization.   

When focusing on the multivariate time-series modality, using knowledge-based abstracted data, rather than the raw-time stamped data, improves the results. The creation of interval-based temporal abstractions of the raw, time-stamped data solves many of the missing value and multiple granularity sampling issues through a form of smoothing. Moreover, combining different levels of abstraction by adding to the pre-processing step Gradient abstraction, which focus on different resolution-level changes in the variables' values, beyond the State abstraction (which focuses on discretization of the values), added considerable prediction power.

Yet another important aspect is the fact that ICD-9 sequences (both as codes or with their textual descriptions) achieve better prediction results than a model that is based solely on the multivariate time-series data (or their abstracted form). A combination between those two only slightly improves the results. Our assumption is that those ICD-9 sequences manage to summarize patients' hospitalization with greater accuracy, focusing on aspects that are more relevant to readmission chances, then the information captured in long sequences of chart items and lab tests.

Finally, our experiments clearly show that a model that is based on discharge notes made be physicians has the best prediction results. Discharge notes are a rich source of unstructured data that contains valuable information about a patient's health status, medical history, and treatment plan. NLP techniques that are geared toward clinical language, such as ClinicalBert, are able to extract key information from these notes and capture complex relationships between clinical variables that might be missed when using structured data alone. Additionally, it is indeed possible that this kind of clinical text can provide context that is not available through structured data, such as patient preferences or social determinants of health, which can further improve the accuracy of readmission predictions. When combining such data with the other data modalities (time-series, ICD-9 sequences) we did not observe any improvement in prediction capabilities.
Overall, the discharge notes were found to be the most valuable resource for predicting patients readmission.

Our next objective is to explore additional approaches for integrating all data modalities, with the aim of enhancing the accuracy of prediction outcomes, and find ways to summarize the multivariate time-series data and ICD-9 sequences in a way that could potentially replace the discharge notes in their role as predictors for readmission, in situations where such data does not exist.

%


%
\bibliographystyle{ACM-Reference-Format}
\bibliography{refs}

%
\appendix
\section{Training details}
\label{sec:training-details}
All models (except the RF model) were trained using a single NVIDIA A-100 GPU. To avoid over-fitting, all models used a stop loss of 7 instead of an epoch limitation. We consider a model to have improved if the AUPRC score for the validation set has risen between two successive evaluations, conducted every 200 steps. If the answer is positive we save the model, otherwise we reduce the learning rates. If the stop loss has been exceeded, we stop the training and return the last saved model. We then use the saved model to evaluate it on the test set. 
We used a batch size of 64. The NLP models use a learning rate of $2\cdot 10^{-5}$, and a reduction rate of 0.9, the rest use $10^{-3}$ and 0.97 respectively.
For each model we perform this training loop for each of the five folds. 

\section{ICU Stays List}
\label{sec:icu-stays}
While the data in MIMIC-III is not available for sharing, it can be accessed online through: \href{https://mimic.mit.edu/}{mimic.mit.edu}. Once access to MIMIC-III has been gained, reproduction of the list of ICU stays can be done using the code found at: \href{https://github.com/meniData/Unplanned-Readmission}{https://github.com/meniData/Unplanned-Readmission}. 


\end{document}